# Apriori-based Analysis of Learned Helplessness in Mathematics Tutoring: Behavioral Patterns by Level, Intervention, and Outcome

**John Paul P. Miranda**

Pampanga State University, Philippines

jppmiranda@pampangastateu.edu.ph

**Abstract**: This study applied the Apriori algorithm to analyze behavioral interaction patterns associated with learned helplessness (LH) in mathematics tutoring system logs. Interaction data were examined across three dimensions: LH level (low vs. high), system-based intervention (with vs. without), and problem-solving outcomes (solved vs. unsolved). The analysis of the complete dataset showed that skipping problems without using hints was the most frequent pattern linked to unsolved outcomes, while persistence behaviors such as not skipping were less dominant overall. Comparisons by LH level showed that low-LH students had stronger links between problem solving and not skipping, as well as positive associations between hint use and solved outcomes. High-LH students showed more avoidance patterns, with skipping strongly tied to unsolved outcomes. In the comparison of system-based intervention conditions, students without intervention had the highest lift for persistence–success links, while the with-intervention group had stronger patterns involving skipping behaviors leading to unsolved outcomes. Outcome-specific analysis showed that not skipping was consistently associated with solved problems across all groups, while skipping without hints predicted unsolved outcomes. Practical implications and recommendations are discussed.

## 1. Introduction

Students' behavior while working on mathematics problems reflects their motivation, persistence, and approach to challenges (Albay, 2020). In some cases, these behaviors indicate learned helplessness (LH), a state in which learners expect failure and reduce their effort (Amadi, Agi, & Nwoke, 2020; Hwang, 2019; Yates, 2009). In mathematics, LH can lead to giving up after mistakes, avoiding difficult problems, or skipping opportunities to seek help (Biber and Biber, 2014; Gürefe and Bakalım, 2018). Intelligent tutoring systems (ITS) have been developed to guide students through practice tasks, track their actions, and offer targeted support. These systems have been shown to help learners by adapting to their needs (Muangprathub, Boonjing, & Chamnongthai, 2020; Spitzer and Moeller, 2023), but they can also reveal patterns of negative behaviors that limit learning gains (Fancsali, 2014; Namukasa et al., 2023; du Plooy, Casteleijn, & Franzsen, 2024; Yang et al., 2022).

The behavioral patterns associated with LH in tutoring systems connect to several motivational frameworks. Attribution theory holds that students who attribute failure to fixed, uncontrollable causes tend to disengage rather than persist (Hwang, 2019; Weiner, 1986). In a tutoring context, this appears as skipping problems after mistakes rather than reattempting them. Self-determination theory adds that when students' need for competence goes unmet, their motivation to engage weakens (Ryan and Deci, 2000). Students who avoid hints may do so partly because prior help-seeking produced no felt sense of progress. Self-regulated learning research treats help-seeking as a deliberate strategy used by effective learners, and links help avoidance to weaker self-regulation and lower achievement (Yang, 2023; Zimmerman, 2000). Skipping and non-use of hints in tutoring logs are therefore not isolated actions because they carry motivational meaning that LH theory can help interpret.

Research on LH has often relied on surveys or experimental tasks, with fewer studies using detailed interaction logs from tutoring systems (Miranda et al., 2025; Yates, 2009). While data mining methods can uncover patterns in large datasets, the Apriori algorithm has frequently been applied to examine how combinations of student behaviors (Bringula et al., 2025; Fu, Ren, & Lin, 2025; Tang et al., 2024) but not in the context of LH particularly in mathematics. There is also limited work comparing these patterns between learners with different LH levels or between those with and without system-based interventions. Without this information, it is difficult to design support features that respond to the specific ways different students engage or disengage.

This study applies the Apriori algorithm to interaction logs from a mathematics tutoring system to identify patterns in student behaviors. Specifically, it aims to: (1) examine the distribution of behavioral indicators such as mistakes, hint use, skipping, and solution status; (2) identify frequent behavioral patterns in the complete dataset; (3) compare these patterns between students with low and high LH levels; (4) compare patterns between students with and without system-based intervention; and (5) determine which patterns are associated with solved and unsolved problems in each group. The results are expected to provide practical insights for improving adaptive features in tutoring systems so they can better support persistence and reduce avoidance behaviors in mathematics learning.

## 2. Methodology

### 2.1 Dataset

The dataset for this study came from AES (Adaptive Equation Sensei) (Fig. 1), a mathematics tutoring system used by Grade 8 students in the Philippines (Miranda and Bringula, 2023; Miranda et al., 2025) based from (Bringula et al., 2015). Data originally came from two separate groups: students who used the system without intervention and those who used it with system-based interventions. The system-based intervention consisted of automated hints triggered by the system when a student made an error, motivational messages displayed during the session, and prompts designed to encourage continued engagement with the problem. Students in the without-intervention group used the same AES system but did not receive these features. Both datasets were stored together after collection for unified analysis. The complete dataset contained 3,696 interaction sessions generated by 246 students, with 113 students in the without-intervention group and 133 in the with-intervention group (Fig. 2). Data were collected in multiple periods between 2024 and early 2025. The initial pool included 193 students from eight public schools in the without-intervention group and 192 students from six different public schools in the with-intervention group.

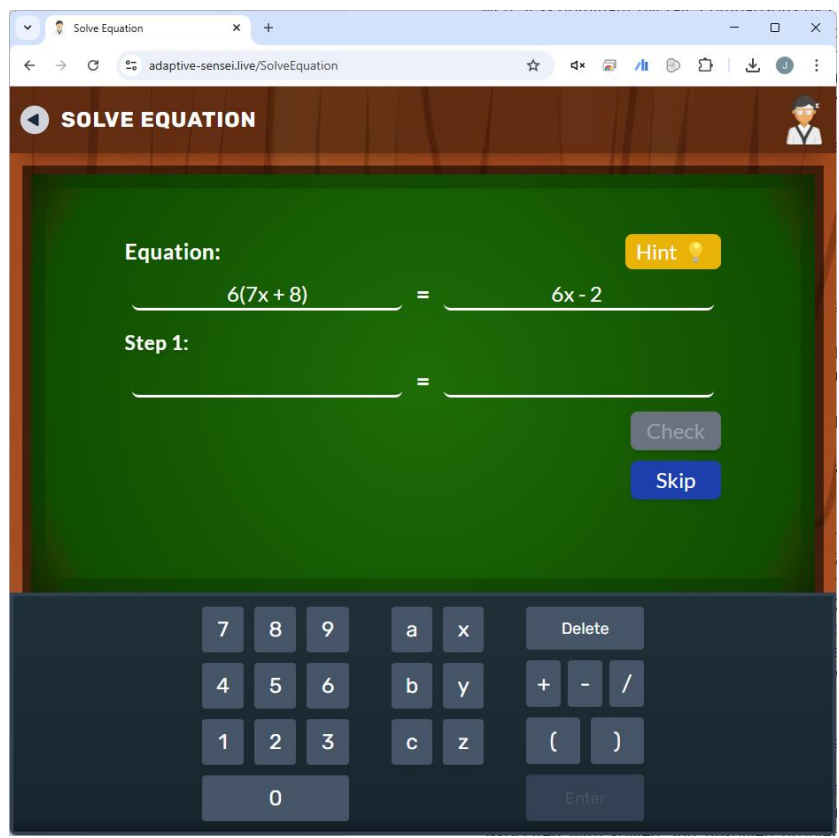

**Figure 1: Adaptive Equation Sensei**

| | A | B | C | D | E | F | G | H | I | J | K |
|---|---|---|---|---|---|---|---|---|---|---|---|
| 1 | Account | MistakeO | HintUsed | Skipped | Status | TotalStep | TotalHints | TotalAnsw | TimeSpen | With Inter | Label |
| 2 | ABIS01 | YES | NO | NO | UNSOLVED | 2 | 0 | 11 | 466 | YES | Low |
| 3 | ABIS01 | NO | NO | NO | UNSOLVED | 2 | 0 | 3 | 120 | YES | Low |
| 4 | ABIS01 | YES | NO | NO | SOLVED | 1 | 0 | 3 | 117 | YES | Low |
| 5 | ABIS01 | YES | NO | YES | UNSOLVED | 2 | 0 | 2 | 238 | YES | Low |
| 6 | ABIS01 | YES | NO | NO | UNSOLVED | 1 | 0 | 3 | 179 | YES | Low |
| 7 | ABIS02 | YES | NO | YES | UNSOLVED | 0 | 0 | 2 | 137 | YES | High |
| 8 | ABIS02 | YES | NO | NO | UNSOLVED | 1 | 0 | 4 | 161 | YES | High |
| 9 | ABIS02 | YES | NO | NO | SOLVED | 1 | 0 | 3 | 91 | YES | High |
| 10 | ABIS02 | NO | NO | NO | SOLVED | 1 | 0 | 1 | 19 | YES | High |
| 11 | ABIS02 | NO | NO | NO | SOLVED | 1 | 0 | 1 | 21 | YES | High |
| 12 | ABIS02 | YES | NO | YES | UNSOLVED | 1 | 0 | 4 | 69 | YES | High |
| 13 | ABIS02 | YES | NO | YES | UNSOLVED | 0 | 0 | 1 | 20 | YES | High |
| 14 | ABIS02 | NO | NO | NO | UNSOLVED | 0 | 0 | 0 | 2 | YES | High |
| 15 | ABIS02 | NO | NO | NO | UNSOLVED | 0 | 0 | 0 | 0 | YES | High |
| 16 | ABIS02 | YES | NO | NO | UNSOLVED | 0 | 0 | 2 | 145 | YES | High |
| 17 | ABIS03 | YES | NO | YES | UNSOLVED | 0 | 0 | 1 | 65 | YES | High |
| 18 | ABIS03 | NO | NO | YES | UNSOLVED | 0 | 0 | 0 | 20 | YES | High |

**Figure 2: Dataset exported to spreadsheet from the mobile tutoring system**

### 2.2 Variables

The analysis used behavioral variables recorded by the tutoring system. For the purposes of this study, a session refers to a single continuous interaction with the AES system by one student, during which one or more problems were attempted, resulting in one row of logged behavioral data. These included binary indicators for mistake occurrence (*MistakeOccurred*), hint use (*HintUsed*), problem skipping (Skipped), and problem status (Status, solved or unsolved). Additional session-level counts included *TotalSteps*, *TotalHints*, and *TotalAnswerAttempts*. Two grouping variables were used: *With Intervention* (with vs. without) and *Label* (low vs. high LH). No demographic variables were included.

The LH label used to classify students into low and high groups was derived from a supervised machine learning classification model developed in prior work using the same AES system (Miranda and Bringula, 2023; Miranda and Bringula, 2025; Miranda et al., 2025). In those studies, LH was operationalized using Yates's (2009) 10-item teacher-rated scale, which served as the ground truth for classifying students into low and high LH groups based on observed behavioral indicators in mathematics. A Random Forest model was then trained on behavioral and academic features extracted from interaction logs, including problem-solving success rates, hint use frequency, and skipping behavior, and evaluated through 10-fold cross-validation, achieving 92% accuracy, a mean F1-score of 0.93, and a recall of 98% for high LH cases (Miranda and Bringula, 2025). In the present study, the binary output of this model was used to assign LH labels to students, allowing behavioral interaction patterns to be compared across low and high LH groups. This approach to operationalizing LH is consistent with the broader use of behavioral indicators as proxies for psychological constructs in educational data mining research, where self-report instruments are not always feasible within system-based data collection contexts (Fancsali, 2014; Yates, 2009).

### 2.3 Data Preparation and Analysis

The dataset was cleaned to remove blank or incomplete records. All retained variables were in binary or categorical form to fit the Apriori algorithm's requirements. Account identifiers were excluded from the pattern mining itself but retained for grouping by intervention status and LH level. Data were filtered to create subgroups for each analysis dimension: LH level, intervention status, and problem-solving outcome. The 3,696 sessions were contributed by 246 students, with an average of approximately 15 sessions per student. Because each session was treated as an independent transaction, repeated observations from the same students are present in the dataset, which has implications for how the results should be interpreted. Within-student dependencies are not accounted for, meaning that behavioral tendencies of individual students are reflected across multiple transactions rather than a single observation. As a result, students with more sessions contribute more heavily to the frequency counts that determine which *itemsets* meet the minimum support threshold. The patterns reported here therefore characterize session-level behavioral co-occurrences across the dataset rather than behavioral profiles of individual students, and findings should be interpreted accordingly rather than as statements about consistent tendencies within a given learner. This approach is consistent with session-level association rule mining in comparable studies (Bringula et al., 2025; Wang, Xiao, & Ma, 2022). Data cleaning and preparation were completed using Python in Jupyter Notebook, with supplementary inspection in spreadsheet software.

The analysis used the Apriori algorithm implemented through the *mlxtend.frequent_patterns* library in Python. The Apriori algorithm was selected over alternative data mining approaches for three reasons. First, it produces association rules expressed through support, confidence, and lift. These metrics are directly readable by educators, school administrators, and tutoring system designers who may not have a background in statistical modelling (Agrawal and Srikant, 1994; Wang, Xiao, & Ma, 2022). Predictive models such as logistic regression or decision trees produce probability estimates that require interpretation within a modelling framework. Apriori, by contrast, produces explicit if-then behavioral rules that can be communicated to practitioners and used directly to inform design decisions. Second, each session in the dataset is represented by a set of binary behavioral indicators. This session-level transactional structure is the format for which Apriori was designed. Third, the research objective is to identify which behavioral indicators co-occur within sessions linked to specific learner profiles and outcomes. It is not to reconstruct the order of individual actions within a session. Sequential pattern mining is suited for detecting ordered event chains in time-stamped logs (De et al., 2022; Real, Pimentel, & Braga, 2021; Zhang and Paquette, 2023). Because this study aggregates behavioral indicators at the session level, temporal ordering is not an analytical concern, and Apriori is the more appropriate method.

Data were first transformed into a transaction format using *TransactionEncoder* from the *mlxtend.preprocessing* module. Minimum thresholds for support, confidence, and lift were set to filter meaningful patterns (support $\geq$ 0.20, confidence $\geq$ 0.60, lift > 1). These thresholds were selected based on conventions established in educational data mining research and on the characteristics of the dataset. A minimum support of 0.20 means that a behavioral pattern must appear in at least 20% of sessions to be considered frequent. This threshold is appropriate for a dataset of 3,696 sessions because it ensures that reported patterns reflect the behavior of a meaningful proportion of the sample, rather than isolated cases (Hikmawati, Maulidevi, & Surendro, 2021; Papadogiannis, Wallace, & Karountzou, 2024). A minimum confidence of 0.60 was applied to retain only rules where the antecedent predicts the consequent in at least 60% of relevant transactions. This level reflects a moderately strong directional association and is consistent with thresholds used in comparable student behavior studies (Bringula et al., 2025; Fu, Ren, & Lin, 2025). A lift threshold greater than 1.0 ensures that the association between two behavioral indicators is stronger than what would be expected by chance alone, which is the minimum condition for a rule to carry practical meaning (Agrawal and Srikant, 1994; Sowan et al., 2025). Only the top 30 rules based on lift values were retained for detailed analysis. Association rules were generated for the entire dataset and separately for subgroups based on LH level, intervention status, and outcome. Descriptive statistics were computed using *pandas* in Python, while the generated *itemsets* and rules were saved to CSV files for review and interpretation.

To examine the stability of the reported patterns, a sensitivity check was conducted by re-running the Apriori algorithm across nine threshold combinations: minimum support values of 0.15, 0.20, and 0.25, crossed with minimum confidence values of 0.50, 0.60, and 0.70, while maintaining a consistent lift threshold greater than 1.0 across all subgroups. As shown in Table 1, the two primary avoidance-related rules, {Skipped} → {Unsolved} and {Skipped, HintUsed = No} → {Unsolved}, returned identical lift values across all nine combinations in every subgroup, indicating that these associations are not artifacts of the specific cutoff values selected for the primary analysis. The persistence-success rule, {Not Skipped} → {Solved}, was detected only in the without-intervention subgroup and only at a minimum confidence of 0.50, which is below the 0.60 threshold used in the primary analysis. The compound rule {Mistake, Skipped} → {Unsolved} was absent in the with-intervention group at a minimum support of 0.25 because its observed itemset support (0.231) fell below that cutoff.

**Table 1: Lift values of key association rules by subgroup across all threshold combinations**

| Subgroup | {Skipped} → {Unsolved} | {Skipped, No Hint} → {Unsolved} | {Mistake, Skipped} → {Unsolved} | {Not Skipped} → {Solved} |
|---|---|---|---|---|
| Overall | 1.244 | 1.244 | 1.244 | - |
| Low LH | 1.261 | 1.261 | 1.261 | - |
| High LH | 1.202 | 1.202 | 1.202 | - |
| Without Intervention | 1.252 | 1.252 | 1.252 | 2.847* |
| With Intervention | 1.231 | 1.231 | 1.231** | - |

*Note*. Lift values are identical across all nine threshold combinations (support ∈ {0.15, 0.20, 0.25}; confidence ∈ {0.50, 0.60, 0.70}; lift > 1.0). A dash indicates the rule did not meet the applicable threshold. * Emerged only at confidence = 0.50. ** Not detected at support = 0.25 (observed support = 0.231).

## 3. Results and Discussion

### 3.1 Distribution of Behavioral Indicators

The behavioral patterns differed between intervention and non-intervention groups, as well as between students with low and high levels of LH. Students in the intervention group were less likely to use hints (85.9%) than those in the non-intervention group (65.8%), while low LH students used hints more often (28.3%) than high LH students (21.1%). Mistake rates were comparable across groups, with high LH students showing a slightly higher frequency (44.4%) than low LH students (41.8%). Skipping behavior was more frequent in the intervention group (53.7%) compared to the non-intervention group (35.1%). This difference may reflect variation in problem-solving approaches rather than a direct effect of the intervention, as the groups consist of different students. Solve rates were generally low across all groups, with the non-intervention group showing a slightly higher proportion of solved problems (20.1%) than the intervention group (18.8%). Across LH levels, low LH students had a higher proportion of solved problems (20.7%) compared to high LH students (16.8%).

### 3.2 Behavioral Patterns Identified Using the Apriori Algorithm

Skipping without using hints was the most frequent pattern associated with unsolved problems (lift = 1.46). Mistakes followed by skipping also appeared frequently, suggesting that errors often led to disengagement. It should be noted that the Apriori algorithm identifies co-occurrence patterns within sessions, not the temporal order of events. Patterns described here as 'mistakes followed by skipping' reflect the co-presence of these behaviors within a session, not a verified sequence in which one event preceded the other. The direction implied by such descriptions is interpretive and should not be taken as evidence of a within-session behavioral chain.

It is also worth noting that skipping a problem does not necessarily indicate avoidance driven by helplessness. In some cases, students may skip because a problem exceeds their current working memory capacity, making it a response to cognitive overload rather than motivational withdrawal (Evans et al., 2024; Kuldas et al., 2014; Nuvvula, 2016; Sweller, 1988; Tsaparlis, 2021). Others may skip strategically to manage limited session time. This behavior is consistent with adaptive self-regulation rather than disengagement (Dang and Koedinger, 2020; Su et al., 2025). The AES system logs record whether a skip occurred but cannot distinguish why. The interpretation of skipping as avoidance in this study rests on its co-occurrence with unsolved outcomes and its association with high LH profiles, not on direct evidence of intent. Future work combining log data with think-aloud protocols or brief post-session surveys could help clarify the motivational basis of this behavior. Patterns linking not skipping with solved problems were present but less common, indicating that avoidance behaviors were more prominent than persistence.

Research supports the role of help-seeking and engagement in shaping learning outcomes. Adaptive help-seeking strategies improve performance, yet many online learners avoid using available assistance (Yang, 2023). Help avoidance, which includes deliberately not using hints, is negatively correlated with learning outcomes and is present in about 19% of students in ITS (Aleven et al., 2006; Li et al., 2024). Students who avoid help often show lower transfer learning scores (Li et al., 2024). Cognitive studies indicate that engaging in challenging activities, such as interleaved practice, enhances memory and problem solving (Samani and Pan, 2021). Motivation also plays a role; higher interest levels are linked to greater persistence and reduced effort avoidance (Song, Kim, & Bong, 2019). These suggest that the frequent skipping and non-use of hints in the dataset reflect effort avoidance and weak self-regulation, while persistence and help-seeking behaviors are associated with better problem-solving performance.

### 3.3 Differences Between Students with Low and High LH Levels

Low LH students showed strong persistence patterns, with solved problems linked to not skipping (lift = 2.33) and a positive link between hint use and solving problems (lift = 1.39). High LH students had stronger associations between skipping and unsolved problems (lift = 1.39), as well as skipping without mistakes and unsolved outcomes (lift = 1.37), reflecting greater avoidance tendencies. Studies on LH indicate that students attributing failure to uncontrollable causes are less persistent and less likely to seek help (Maier and Seligman, 2016). Help avoidance and ineffective strategies such as "wheel spinning" can reinforce feelings of helplessness (Beck and Gong, 2013; Li et al., 2024; Sideridis, 2003; Song, Kim, & Bong, 2019). In contrast, adaptive help-seeking is linked to improved performance and better coping with challenges (Li, Che Hassan, & Saharuddin, 2023). The patterns observed suggest that low LH students benefit from persistence and help use, while high LH students' avoidance behaviors correspond with unsolved outcomes.

### 3.4 Differences Between Students with and Without Intervention

Students without intervention exhibited the strongest persistence-success association, with solved problems linked to not skipping (lift = 2.85). They also showed a notable pattern connecting hint use to mistakes (lift = 1.77). In the intervention group, skipping was closely tied to unsolved problems (lift = 1.35), and mistakes in unsolved problems often preceded skipping (lift = 1.34). Research on ITS shows that feedback on help-seeking can encourage hint use but may not lead to higher achievement (Aleven et al., 2016). Although hints can reduce floundering, some learners still avoid them (Aleven et al., 2016; Borchers et al., 2025; Li et al., 2024). Success in online learning requires self-regulation and effective help-seeking strategies, yet interventions alone may not overcome avoidance behaviors (Yang, 2023). These differences, however, should be interpreted with caution. The two groups consisted of different students from different schools, and the absence of randomization means that observed differences in behavioral patterns may reflect pre-existing group characteristics rather than the effect of the intervention itself.

### 3.5 Patterns Strongly Associated with Solved and Unsolved Outcomes for Each Group

In all groups, not skipping was positively associated with solved problems, with the strongest lifts in the without-intervention group (lift = 2.85) and among low LH students (lift = 1.54). Hint use was also positively associated with solved problems in the low LH group (lift = 1.39). Skipping without hints was consistently linked to unsolved outcomes, particularly in the low LH (lift = 1.50) and high LH (lift = 1.48) groups. Both intervention conditions showed skipping behaviors as defining features of unsolved problems. Empirical evidence confirms that help avoidance is a predictor of poor outcomes, while help seeking and persistence foster academic success. Students avoiding assistance show lower performance (Aleven et al., 2006; Li et al., 2024), whereas those who actively seek help cope better with challenges (Li, Che Hassan, & Saharuddin, 2023). Higher interest reduces effort avoidance and promotes persistence (Song, Kim, & Bong, 2019). Even though difficult learning strategies such as interleaved practice are perceived as harder, they lead to better problem-solving ability (Samani and Pan, 2021). Not skipping and using hints are therefore associated with effective self-regulation, while skipping without hints is a consistent marker of unsolved outcomes across all groups.

## 4. Conclusion and Future Work

This study applied the Apriori algorithm to interaction logs from a mathematics tutoring system to examine behavioral patterns related to LH across differences in LH level, intervention condition, and problem-solving outcome. The analysis found that not using available hints was the most frequent pattern linked to unsolved problems, while persistence behaviors, such as continuing with a problem, appeared less often. Low-LH students showed stronger links between persistence and solved problems, as well as positive associations between hint use and correct solutions. High-LH students displayed more avoidance-oriented patterns, particularly skipping behaviors tied to unsolved results. Students without system-based intervention exhibited stronger persistence-success associations, while those with intervention tended to show more skipping behaviors. Across all groups, continued engagement with problems was consistently related to solved outcomes, whereas avoidance without hint use was more often connected to unsolved ones.

There are several limitations in this study. The dataset did not include demographic or contextual variables that might influence engagement patterns. The LH classification was derived from a Random Forest model trained on behavioral features and grounded in Yates' (2009) validated teacher-rated scale; nonetheless, the use of a model-derived binary label as a proxy for a psychological construct introduces some degree of construct validity limitation in how the low and high LH groupings should be interpreted. The intervention and non-intervention groups were drawn from different schools without randomization, so differences in behavioral patterns between groups cannot be attributed to the intervention and should be interpreted as associations rather than causal effects. The study focused on Grade 8 learners in the Philippines, so patterns may differ in other educational levels or contexts. The session-level unit of analysis does not account for within-student dependencies, as individual students contributed multiple sessions to the dataset. This means that behavioral tendencies of high-frequency contributors are reflected more prominently in the reported patterns, and the findings should be interpreted as characterizing session-level co-occurrences rather than consistent behavioral tendencies within individual learners. Future studies employing student-level aggregation or multilevel modeling approaches could more precisely examine whether the reported associations generalize across individual students. Because behavioral indicators were aggregated at the session level, within-session event sequences were not modeled, and future studies with timestamped action-level logs are encouraged to apply other pattern mining alongside association rule mining. A sensitivity check confirmed that the primary avoidance-related patterns were stable across varied threshold combinations, though future work may explore a wider range of values to further establish robustness. The analysis also relied solely on log data, which captures observable behaviors but not the underlying cognitive or emotional processes that drive them.

Despite these constraints, the observed patterns suggest several opportunities for improving ITS. Adaptive features could be designed to detect early signs of avoidance, such as repeated skipping or low hint use, and to provide timely prompts that encourage persistence and improve help-seeking. For students with higher LH, strategies may involve guiding them toward productive hint use and reinforcing persistence after mistakes. For students with lower LH, approaches could aim to sustain engagement and gradually increase problem difficulty to strengthen resilience. Educators might also use these patterns to identify students who could benefit from targeted support, combining system data with classroom observations for a more complete understanding of learning behaviors. Future enhancements to the AES platform may include real-time pattern detection, dynamic adjustment of problem difficulty, and context-sensitive feedback. Additional studies that integrate behavioral logs with self-reported or interview data could help explain the motivational and emotional factors driving these behaviors.

**Ethical Statement**: This study used student data collected by the author under ethics clearance from the UE Ethics Review Committee, with informed consent obtained from all participants and full compliance with the Data Privacy Act of 2012.

**AI Ethics Statement**: The author confirm that they did not use AI when writing this study.

*John Paul P. Miranda*
Spitzer, Markus Wolfgang Hermann and Moeller, Korbinian. (2023). Performance increases in mathematics during COVID-19 pandemic distance learning in Austria: Evidence from an intelligent tutoring system for mathematics. *Trends in Neuroscience and Education*, 31, pp. 100203. doi:https://doi.org/10.1016/j.tine.2023.100203

Su, Man ; Dang, Belle ; Nguyen, Andy and Nagashima, Tomohiro. (2025). Choice-making in an adaptive learning system with motivational pedagogical agents. *NPJ Science of Learning*, 10(1), pp. 77. doi:10.1038/s41539-025-00366-7

Sweller, John. (1988). Cognitive Load During Problem Solving: Effects on Learning. *Cognitive Science*, 12(2), pp. 257–285. doi:https://doi.org/10.1207/s15516709cog1202_4

Tang, Zheng ; Jiang, Zhengwei ; Li, Ying ; Yuan, Haowei ; Han, Jiayu and Chen, Chao. (2024). Research on the Association Analysis of Online Learning Behaviors Based on the Apriori Algorithm. *Frontiers in Computing and Intelligent Systems*, 9(2), pp. 18–22. doi:10.54097/4a3h3p03

Tsaparlis, Georgios. (2021). Chapter 5. It Depends on the Problem and on the Solver: An Overview of the Working Memory Overload Hypothesis, Its Applicability and Its Limitations. In *The Royal Society of Chemistry*. doi:https://doi.org/10.1039/9781839163586-00093

Wang, Tengfei ; Xiao, Baorong and Ma, Weixiao. (2022). Student Behavior Data Analysis Based on Association Rule Mining. *International Journal of Computational Intelligence Systems*, 15(1), pp. 32. doi:10.1007/s44196-022-00087-4

Weiner, Bernard. (1986). *An Attributional Theory of Motivation and Emotion* (1st edn). Springer New York, NY. doi:https://doi.org/10.1007/978-1-4612-4948-1

Yang, Albert C M ; Chen, Irene Y L ; Flanagan, Brendan and Ogata, Hiroaki. (2022). How students' self-assessment behavior affects their online learning performance. *Computers and Education: Artificial Intelligence*, 3, pp. 100058. doi:https://doi.org/10.1016/j.caeai.2022.100058

Yang, Fan. (2023). A Systematic Review of Studies Exploring Help-Seeking Strategies in Online Learning Environments. *Online Learning Journal*, 27(1), pp. 107–126. doi:10.24059/olj.v27i1.3400

Yates, Shirley. (2009). Teacher Identification of Student Learned Helplessness in Mathematics. *Mathematics Education Research Journal*, 21(3), pp. 86–106.

Zhang, Yingbin and Paquette, Luc. (2023). Sequential Pattern Mining in Educational Data: The Application Context, Potential, Strengths, and Limitations. In Alejandro Peña-Ayala (Ed.). *Educational Data Science: Essentials, Approaches, and Tendencies: Proactive Education based on Empirical Big Data Evidence*. Singapore : Springer Nature Singapore. doi:10.1007/978-981-99-0026-8_6

Zimmerman, Barry J. (2000). Self-Efficacy: An Essential Motive to Learn. *Contemporary Educational Psychology*, 25(1), pp. 82–91. doi: https://doi.org/10.1006/ceps.1999.1016